%% file: main.tex
\documentclass{article}

\usepackage[T1]{fontenc}
\usepackage{graphicx}
\usepackage{booktabs}
\usepackage[misc]{ifsym}

\usepackage{amsmath}
\usepackage{amsfonts}
\usepackage{hyperref}
\usepackage{xcolor}
\usepackage{ulem}
\usepackage{snaptodo}

\snaptodoset{block rise=2em}
\snaptodoset{margin block/.style={font=\tiny}}

\DeclareMathOperator*{\argmax}{argmax}
\usepackage[final,nonatbib]{style}

\usepackage[utf8]{inputenc} 
\usepackage[T1]{fontenc}    
\usepackage{hyperref}       
\usepackage{url}            
\usepackage{booktabs}       
\usepackage{amsfonts}       
\usepackage{nicefrac}       
\usepackage{microtype}      
\usepackage{xcolor}         

\title{Better Capturing Interactions between Products in Retail: Revisited Negative Sampling for Basket Choice Modeling}

\author{%
Jules D\'esir$^{1}$ \quad Vincent Auriau$^{1, 2}$ \quad Martin Mo\v{z}ina$^3$ \quad Emmanuel Malherbe$^1$ \\
$^1$Artefact Research Center, Paris, France\\
$^2$MICS, CentraleSup\'elec Universit\'e Paris-Saclay, Gif-sur-Yvette, France \\ $^3$Fortenova Group, Zagreb, Croatia\\
\texttt{\{jules.désir,vincent.auriau,emmanuel.malherbe\}@artefact.com}\\
\texttt{martin.mozina@mercator.si}
}

\begin{document}

\maketitle

\begin{abstract}

Brick-and-mortar retailers face many different challenges that involve understanding thoroughly its products catalog and customer preferences. In particular, assortment optimization - proposing the ideal mix of products - and promotion planning hold a pivotal role in their strategy. By leveraging sales data, retailers can make informed decisions on which products to sell and how to manage inventory, based on customer preferences as well as regional and seasonal trends. It is especially crucial to capture interactions between products, in order to minimize the number of items that cannibalize each other’s sales and to ensure that complementary products, which are often purchased together, are conjointly available and sold across all stores.
In this paper, we propose a model of shopping basket that learns embeddings to represent interactions between products, prices and stores. Our model is built to uncover sales patterns from very large transaction datasets. In particular, the optimization loss is computed with random negative samples in order to overcome the computational bottlenecks that arise with large number of items.
Our experiments on synthetic data show the efficiency of drawing such negative samples based on the actual assortment of available products, with better results than approaches from the literature.
We also validate our approach by training and evaluating our model on a dataset composed of millions of transactions from \textit{Fortenova Group}, one of the leading retail company in Southeast Europe. Our model showcases promising applications in the sector of retail, with enriched interfaces to efficiently support category managers.
\end{abstract}

\section{Introduction}

Choosing the optimal combination of products to sell at the right price is a crucial stake for retailers. It is needed to select which products, among the hundreds of thousands offered by various suppliers, should be offered for sale in each store, or should be under promotion. This decision can be taken using insights such as past revenues or current trends. 

It is also highly crucial to take into consideration the different products synergies. Indeed, a customer planning to bake a cake should be able to purchase all the necessary ingredients or he may choose to shop at another supermarket. This notion of complementarity between products should be carefully considered, along with cannibalization, where the presence of two similar items lead to a reduction in each other's sales. Historically, this assortment selection has been done manually by the category managers, based on their business experience. With thousands of daily customers in its different stores, a supermarket company typically records hundreds of millions of purchases per year. 
This data - baskets of co-purchased items - conveys rich patterns, especially compared to assortments which remain relatively stable due to logistical reasons.
Gathering statistical insights from these data can greatly help category managers make informed decisions on the assortment selection.

Choice modeling is a common approach that leverages purchase datasets to understand customers preferences. Each product is assigned a utility, a latent value that reflects its probability to be purchased. Common models use a linear representation of utility \cite{mcfadden1974conditional-logit,RUMMcFadden}, while more recent work has shown that incorporating non-linear effects can help in better capturing the heterogeneity in the different customer behaviors. In particular, neural networks \cite{aouad2022rumnet,arkoudi2023ann-based} have shown promising performances. Besides, estimating and leveraging product interactions is a problem yet to be tackled.
Indeed, the exponentially growing number of interactions to take into account leads to difficulties in using classical methods \cite{sobolewski2017estimating, basket2022caputo} with a large number of products.
Another approach, the nested logit model \cite{nested}, requires the modeler to decide a priori the sets of substitutable items, which is on the contrary what we want the model to learn. 

In this paper, we propose a neural-network choice model that learns latent representations of items, shops and baskets, with a structure capable of capturing product interactions from a large dataset of purchase receipts. It enables to interpret the choice probabilities predicted by the model, and has promising applications for enriching interfaces used by category managers. Such interfaces are used to plan promotion campaigns or select the assortment of a shop and would be more informative if the interactions between products were displayed.

Our contributions in this paper are the following.
First, we propose a choice model capturing interactions between products, able to be learned on large-scale data with no or few features to describe items, shops or customers. Second, we propose to leverage the assortment of available products in a soft manner within the negative sampling during the training procedure. Last, we present experiments on synthetic data and a real-world retailer dataset, confirming the benefits of our approach when compared to existing models. Our code is publicly available on GitHub\footnote[1]{The different Python implementations can be found in the Choice-Learn package at \url{https://github.com/artefactory/choice-learn}.}.

\section{Related Work}

Our paper relates to two strands of the literature: discrete choice models and representation learning, with similarities with word embeddings.

\subsection{Discrete Choice Models}

Discrete choice models try to explain or predict choice decisions made by humans from a set of alternatives, called an assortment. They have been widely adopted for a large range of applications, usually involving a customer decision, such as choice of transportation mode or products purchases.
The discrete choice framework based on Random Utility Maximization \cite{RUMMcFadden} considers that an individual attributes a utility to the different options and chooses the one that maximizes this utility. The objective of discrete choice modeling is to estimate this latent utility function harnessing data with past choices or preferences. The classical conditional logit model \cite{mcfadden1974conditional-logit} formulates a linear utility function leveraging the different available features of the problem. Recent works use more complex functions that introduce nonlinearities, using for example neural networks 
\cite{wang2020asu-dnn,han2022tastenet,aouad2022rumnet,sifringer2018enhancing} or tree-based models \cite{salvade2025rumboost,aouad2023market}.

Discrete choice models are designed to handle the variations of options availability within the assortment. This is particularly important in a retail context where products might be out of stock or simply not sold in specific locations. However, they are usually built to model a single choice and still face challenges when modeling bundles of options, like we observe in retail. These challenges result in limited methods for estimating interactions between these choices. Interaction terms can be introduced within the utility function \cite{sobolewski2017estimating,basket2022caputo}, but such formulations are typically restricted to a small number of distinct options. Indeed, the number of possible bundles grows exponentially with the number of available items, which can be very high in applications such as retail.

\subsection{Representation learning}

Object representation in machine learning has significantly evolved  with the advent of neural networks \cite{lecun2015deep}. Unlike traditional machine learning methods, which could mainly transform pre-existing features, neural networks can learn continuous representations when the data is categorical and comes with no features \cite{guo2016entity}. Traditionally, such categorical data is represented using one-hot vectors \cite{seger2018investigation} that have the size of the set of possible values. In comparison, learned embeddings provide a continuous representation that replaces these vectors in a generally much lower dimension. One may note that the term "embedding" sometimes refers to a transformed object's features after multiple operations \cite{gong2014multi}, but in this paper it will only refer to categorical data representation.

One of the primary data structures that have greatly benefited from embeddings is word representation \cite{pilehvar2020embeddings}. Traditionally, words were represented using sparse vectors \cite{liu2018research}, similar to weighted one-hot vectors. \cite{mikolov2013distributed-rep-words} proved the efficiency to train a neural network where each word in the vocabulary is associated with a learned dense continuous vector. This training process typically requires large-scale data and is often conducted in a self-supervised manner using natural data \cite{balestriero2023cookbook}, i.e. an unlabeled corpus of documents. In this case, negative sampling \cite{yang2024does} is commonly employed to complement the positive cases derived from the natural data. This involves randomly sampling words from the vocabulary, under the assumption that they are likely to be implausible outputs.

Embeddings have also found applications beyond words. They can be used for any categorical data, typically the labels of a multiclass classification \cite{rodriguez2018beyond,akata2015label}, the nodes of a graph \cite{xu2021understanding}, or the rows of tabular data represented as graphs \cite{kim2024carte}. In a scenario similar to choice modeling, collaborative filtering for recommender systems, matrix factorization or neural networks aim at building embeddings for each user and product \cite{koren2021advances,li2024embedding}. In this case, the model is generally trained on the scores given by users on products, such as movies \cite{gomez2015netflix}.
To our knowledge, limited research has proposed embedding structures for neural network-based choice models. Among existing work, \cite{arkoudi2023ann-based} and \cite{pereira2019rethinking} embeddings are fed directly into the softmax function and thus do not model any interaction. \cite{ko2023modeling} incorporates interaction terms with respect to the items in the assortments, but focuses solely on single-choice modeling, which does not fully leverage the rich information provided by bundles of items purchased together. \cite{ruiz2020shopper} presents a closer approach to our work; however, their method requires heavy computing since their  neural network is Bayesian and they need to generate all subsets of items within a bundle, both during training and inference.

\section{Model}

\subsection{Context and Objectives}
\label{sec:objectives}
We build a model with the following characteristics in mind:
\begin{itemize}
\item Only a few or no features are available to characterize the items other than their price and whether they are on promotion. The main signals come from co-purchase patterns. 
\item The model should capture complementarity and cannibalization between products — i.e. respectively the tendency for items to be purchased together within a single basket and the reduction of sales when items are both available in the assortment.
\item The model is designed to efficiently handle large datasets, both in terms of the number of items considered as well as the volume of purchases.
\end{itemize}
This setup corresponds to our retail scenario and would apply to other sectors with brick-and-mortar stores.

\subsection{Notations}
\label{sec:notations}

The data we consider consists of a collection of purchases, each representing a basket of items bought by a customer. Additional contextual information includes the store where the purchases took place, the list of items that were available at that time, and their prices. More formally, we write $\mathcal{I}$ the set of items sold in the set $\mathcal{S}$ of stores.
The observed purchases correspond to a tuple $(\mathcal{B}, \mathcal{A}, s, p)$ with:
\begin{itemize}
  \item $\mathcal{B} \subseteq \mathcal{I}$ the basket of purchased items;
  \item $\mathcal{A} \subseteq \mathcal{I}$ the products assortment available at the time of the purchase (by definition, $\mathcal{B} \subseteq \mathcal{A}$);
  \item $s \in \mathcal{S}$ the store in which the purchase happened;
  \item $p$ the prices for all the items in $\mathcal{A}$ at the time of the purchase. In particular, $p_i$ is the price of item $i$.
\end{itemize}
We decide not to consider duplicated items in the basket. This choice reflects our objective of capturing interactions between items of the catalog.

\subsection{Conditional definition of the utility}

We propose a conditional formulation of the utility. Considering an already chosen basket of items $\mathcal{B}$, we write the utility of an additional item as $V(i | \mathcal{B}, s, p)$, modeled as two independent contributions:
\begin{equation}
    V(i | \mathcal{B}, s, p) = V^{pref}(i | s, p) + V^{inter}(i | \mathcal{B})
\end{equation}
where $V^{pref}$ represents the intrinsic utility (or "preference") attributed to the item and the interaction utility $V^{inter}$ depends on the other items that will also be purchased. $V^{pref}$ can be understood as the popularity of an item, regardless of the other items for sale. In our case, it would only depend on the item, its price and the store where it is purchased. 

Following classical formulations of the utility, we define $V^{pref}$ as an itemwise intercept, a logarithmic contribution of the price and an elasticity that depends on the item and the store:
\begin{equation}
    V^{pref}(i | s, p) = \alpha_i + \boldsymbol{\beta}_i \cdot \boldsymbol{\delta}_s \log p_{i}
\end{equation}
where $\cdot$ denotes the scalar product and variables in bold are vectors. We have $\alpha_i \in \mathbb{R}$ and $\boldsymbol{\beta}_i, \boldsymbol{\delta}_s \in \mathbb{R}^K$, with $K \in \mathbb{N}^{*}$ the size of the latent representation to be chosen. $\boldsymbol{\delta}_s$ captures that each store has a different population with its own elasticities. 
 The formulation of $V^{pref}(i | s, p)$ can be easily modified to introduce multiple underlying factors governing customer choice.

For the interaction term, we use a formulation  inspired by methods such as Word2Vec \cite{mikolov2013distributed-rep-words}, where, in a similar fashion, the probability that a word matches its surrounding words is estimated. In our case, the context is given by items in the basket $\mathcal{B}$, so that $V^{inter}$ is expressed as:
\begin{equation}
    V^{inter}(i | \mathcal{B}) = \boldsymbol{\gamma_i} \cdot \boldsymbol{E}_{\mathcal{B}}
\end{equation}
with $\gamma_i \in \mathbb{R}^K$ a learned representation of item $i$ and $\boldsymbol{E}_{\mathcal{B}}$ the representation for basket $\mathcal{B}$. Different formulations of such basket embedding could be used, we have opted for: $\boldsymbol{E}_{\mathcal{B}} = \frac{1}{|\mathcal{B}|} \sum_{j \in \mathcal{B}} \boldsymbol{\gamma_j}$. One notes that it would be possible to learn different items embeddings for $\boldsymbol{E}_{\mathcal{B}}$ than the $\boldsymbol{\gamma_j}$ ("encoding-decoding" structure).

We assume that the customer behaves rationally and will maximize their utility during a shopping session. Following Discrete Choice Models formulation, we consider that the final utility $U$ is the sum of the deterministic part $V$ as well as a random contribution, $\epsilon_i$:
$$U(i | \mathcal{B}, s, p) = V(i | \mathcal{B}, s, p) + \epsilon_i$$
from which we can derive probabilities, as described in the following section.

\subsection{Purchase probability}
The random variable $\epsilon_{i}$ is assumed to be independent across items and to follow a zero-mean Gumbel distribution (generalized type I extreme value). Consequently, the conditional choice probability of choosing an item $i \in \mathcal{A}$ that is not yet in the basket is a softmax over the remaining items in the assortment $\mathcal{A}$:

\begin{align}
    \mathbb{P}(i | \mathcal{B}, s, p) &= 
    \mathbb{P}\Big(U\left(i| \mathcal{B}, s, p\right) > U\left(j| \mathcal{B}, s, p\right) \,\, \forall j \in \mathcal{A}  \setminus \mathcal{B}, j \neq i \Big)
    \\
    &=
    \frac{e^{V(i | \mathcal{B}, s, p)}}{\sum_{j \in (\mathcal{A} \setminus \mathcal{B})}e^{V(j | \mathcal{B}, s, p)}} 
    \label{eq:proba}
\end{align}
where $\mathbb{P}(i | \mathcal{B}, s, p)$ is the probability to add the item $i$ to the basket  $\mathcal{B}$ at the store $s$ with the prices $p$ and is estimated with the latent parameters $\Theta = \{\alpha, \boldsymbol{\beta}, \boldsymbol{\delta}, \boldsymbol{\gamma}\}$.
We set this probability to zero when $i \notin \mathcal{A}$ or $i \in \mathcal{B}$.

\section{Implementation for large-scale data}
Given a large dataset of $N$ shopping sessions $\{(\mathcal{B}_k, \mathcal{A}_k, s_k, p_k)\}_{k=1}^{N}$, our goal is to infer the latent parameters $\Theta \subset \{\alpha, \boldsymbol{\beta}, \boldsymbol{\delta}, \boldsymbol{\gamma}\}$ which are the trainable weights of our model. 
We now describe how to learn the optimal $\Theta^*$ from the dataset.

\subsection{Training Procedure}

The model parameters $\Theta^*$ are estimated by leveraging all baskets data. To this end, similarly to \cite{basket2022caputo} and \cite{molenberghs2005models} (Chapter 12), we aim at maximizing the quantity:

$$\Theta^* = \argmax_\Theta
\prod_{k=1}^N \prod_{i\in \mathbb{B}_k} \mathbb{P}(i | \mathcal{B}_k \backslash i, s_k, p_k)$$
which can be seen as a form of self-supervision similar to text, where the missing word is predicted based on the remaining words of its context sentence. One notes that it differs from \cite{ruiz2020shopper} where they maximize the probability of the baskets, which is not explicitly defined in our model. Besides, for this purpose they generate all possible sub-baskets $\mathcal{B}' \subseteq \mathcal{B}$, which we believe is not always informative and makes computation exponentially costly as the basket size grows.

One also notes that the formula of $\mathbb{P}(i | \mathcal{B}_k\backslash i, s_k, p_k)$ (Equation \ref{eq:proba}) is computationally heavy too, since $\mathcal{A}$ can be very large. Instead, we normalize $e^{V(i | \mathcal{B}, s, p)}$ over a set of randomly drawn negative samples, formally written $\mathcal{NS}$ (whose size $|\mathcal{NS}|$ is a hyperparameter).
We thus approximate $\mathbb{P}(i | \mathcal{B}_k\backslash i, s_k, p_k)$ with:

$$f(i,\mathcal{B}_k\backslash i, s_k, p_k)= \frac{e^{V(i | \mathcal{B}_k, s_k, p_k)}}{e^{V(i | \mathcal{B}_k, s_k, p_k)} + \sum_{j \in \mathcal{NS}} e^{V(j | \mathcal{B}_k, s_k, p_k)}}$$

Such negative sampling has also been used in \cite{gutmann2012noise} and \cite{mikolov2013distributed-rep-words}, and can be shown to approximate well the log probabilities of the softmax. This strategy is particularly important in a large-scale environment with thousands of different products.

In practice, we maximize a lower bound of
$\prod_{k=1}^N \prod_{i\in \mathcal{B}_k} f(i | \mathcal{B}_k\backslash i, s_k, p_k)$:

$$
\prod_{k=1}^N \prod_{i \in \mathcal{B}_k}
f(i | \mathcal{B}_k\backslash i, s_k, p_k)
\geq
\prod_{k=1}^N \prod_{i \in \mathcal{B}_k}
\frac{e^{V(i | \mathcal{B}_k, s_k, p_k)}}{ (1 + e^{V(i | \mathcal{B}_k, s_k, p_k)})
\prod_{j \in \mathcal{NS}}
(1 + e^{V(j | \mathcal{B}_k, s_k, p_k)})}
$$
which gives us the final loss to be minimized during our training, that is the opposite of the log of the lower bound:
\begin{align}
    \mathcal{L} = - \sum_{k=1}^N \sum_{i \in \mathcal{B}_k} \left( \log \sigma [V(i | \mathcal{B}_k\backslash i, s_k, p_k)] + \sum_{j \in \mathcal{NS}} \log \left(1 -\sigma \left[V\left(j | \mathcal{B}_k\backslash i, s_k, p_k\right)\right] \right) \right)
    \label{eq:loss}
\end{align}
where $\sigma[x]=\frac{e^x}{1+e^x}$ is the sigmoid function.
One notes that this loss corresponds to a binary cross-entropy where the $V(i|\mathcal{B}_k\backslash i, s_k, p_k)$ are associated to the positive labels and the negative samples $V(j|\mathcal{B}_k\backslash i, s_k, p_k)$ to the negative labels.

This loss is minimized during a stochastic gradient descent. At each batch of baskets, the negative samples $\mathcal{NS}$ are randomly drawn, with a different sampling for each epoch in the case of multiple epochs on the dataset. In the following, we describe our method to randomly sample $\mathcal{NS}$.

\subsection{Negative Sampling}

In scenarios involving a fixed set of options, such as the vocabulary in the context of natural language \cite{mikolov2013distributed-rep-words}, negative samples $\mathcal{NS}$ are uniformly sampled from all other options, excluding the true option. In our choice modeling scenario, however, there are several key differences. First, we exclude items already present in basket $\mathcal{B}_k$, as our context does not allow for duplicated items (see Sections \ref{sec:objectives} and \ref{sec:notations}).
Second, we propose to draw negative samples $\mathcal{NS}$ exclusively from items within the available assortment $\mathcal{A}_k$. Formally, we enforce the condition: 
\begin{equation}
    \mathcal{NS} \subset \mathcal{A}_k \backslash \mathcal{B}_k
    \label{eq:NS}
\end{equation}
Implicitly, the negative samples are generated depending on both 
$\mathcal{A}_k$ and 
$\mathcal{B}_k$ on the data point $k$. It could be explicitly written $\mathcal{NS}(\mathcal{A}_k,\mathcal{B}_k)$. Practically, it means that $\mathcal{A}_k$ must be provided as information during the training process.
In contrast, \cite{ruiz2020shopper} draws their negative samples from the entire item catalog outside the considered basket, that is to say $\mathcal{I}\backslash \mathcal{B}_k$.

The rationale for the use of the assortment during negative sampling is that it prevents the model from considering that two items are cannibalizing each other just because they are not available both at the same time in the training data. Indeed, when item $j \in \mathcal{I}\backslash \mathcal{B}_k$ is drawn as a negative sample of item $i \in \mathcal{B}_k$, then the conditional utility $V\left(j | \mathcal{B}_k\backslash i, s_k, p_k\right)$ will be reduced during the gradient descent that minimizes the loss $\mathcal{L}$ (see Equation \ref{eq:loss}). Imposing item $j$ to be in the assortment $\mathcal{A}_k$ means that such reduction of a conditional probability during the gradient descent will happen only to items that could have been bought with item $i$ but had not, leaving items outside of the assortment untouched and thus enabling the model to learn a difference between items that are in the assortment and items that are not. In other words, drawing a negative sample $j \in \mathcal{I}\backslash \mathcal{B}_k $ that is not in assortment $\mathcal{A}_k$ amounts to assuming that $j$ would have been cannibalizing item $i$ if $j$ were in assortment $\mathcal{A}_k$, which is not necessarily true.

We consider this method of incorporating assortment $\mathcal{A}_k$ in the estimation of $\mathbb{P}(i | \mathcal{B}_k\backslash i, s_{k}, p_{k})$ to be simple yet sufficient and effective, with significant benefits, as demonstrated in the following section.

\section{Experiments}

We have conducted several types of experiments that validate the relevance of our approach. First, synthetic experiments let us showcase how it behaves on a smaller and controlled environment. Second, we share results obtained on a large real-world industrial dataset issued from Fortenova Group's supermarkets.

\subsection{Synthetic Experiments}

\label{sec:synthetic}
\subsubsection{Dataset}

\begin{figure}[h]
\centerline{\includegraphics[width=1\textwidth, keepaspectratio]{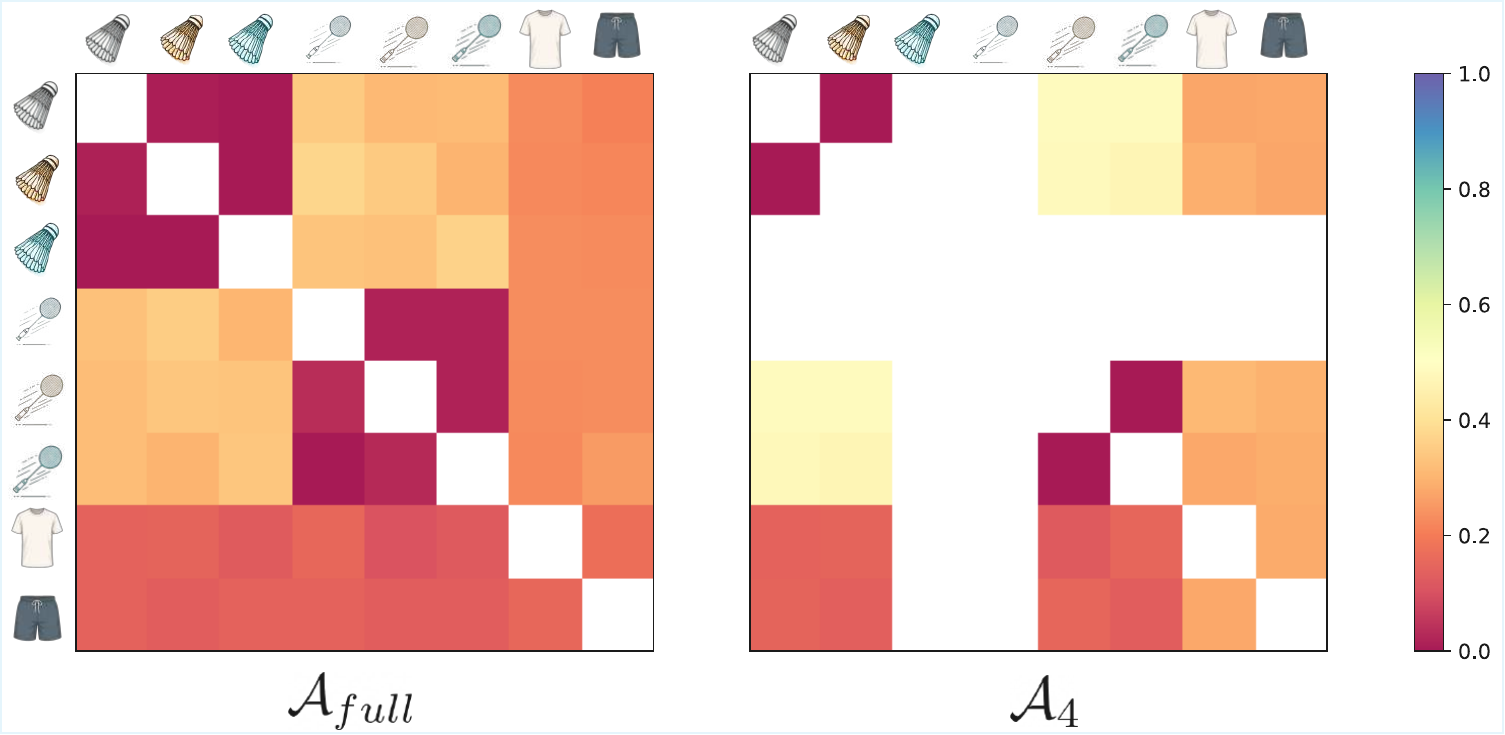}}
\caption{$\mathbb{P}(i \in \mathcal{B}|j \in \mathcal{B})$ for all item pairs $(i, j)$, computed on the synthetic baskets $\mathcal{D}_{full}$ on the left and $\mathcal{D}_{4}$ on the right. Low values correspond to cannibalization and high values to complementarity. }\label{fig:interaction_matrix}
\end{figure}

\begin{figure}
    \centering
    \includegraphics[width=1.0\linewidth]{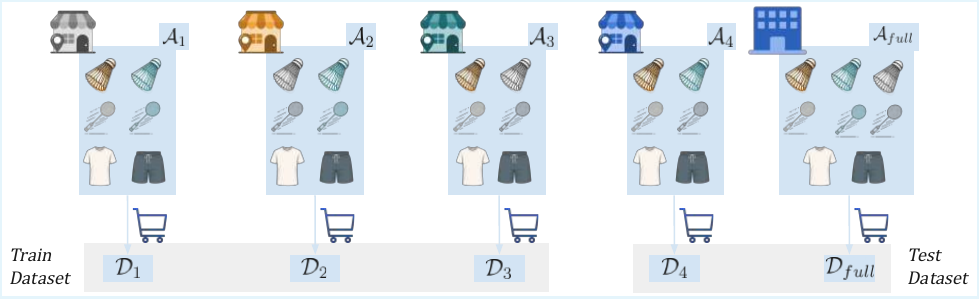}
    \caption{Illustration of the synthetic data assortments. The full set of items $\mathcal{I}$ is in $\mathcal{A}_{full}$.}
    \label{fig:synth_data}
\end{figure}

We build a synthetic dataset for which the catalog has 8 items, $\mathcal{I}=\{1, \ldots,  8\}$, with the following interactions:
\begin{itemize}
    \item \textbf{Cannibalization:} $\{1, 2, 3\}$ on the one hand and $\{4, 5, 6\}$ on the other hand form groups of items cannibalizing each other.
    \item \textbf{Complementarity:} each of the items in $\{1, 2, 3\}$ are complementary to each of the items in  $\{4, 5, 6\}$;
    \item \textbf{Neutral:} $7$ and $8$ are neutral in the sense that they don't have specific interaction with other items.
\end{itemize}
As a concrete example, we can consider that items $\{1, 2, 3\}$ are badminton shuttlecocks,  items $\{4, 5, 6\}$ are badminton rackets while items $\{7, 8\}$ are a t-shirt and a short.
We implemented a routine that, given an assortment $\mathcal{A} \subseteq \mathcal{I}$, generates a dataset $\mathcal{D}$ of purchased baskets. These baskets are generated following the interactions described above, without price or shop effect (apart from its assortment). The details of this implementation are in Appendix 
A\footnote[2]{Appendix can be found in the GitHub repository at \url{https://github.com/artefactory/alea-carta-est}}.
In the left of Figure \ref{fig:interaction_matrix}, we illustrate the simulated interactions by computing $\mathbb{P}(i \in \mathcal{B}|j \in \mathcal{B})$ on baskets generated by our routine for a full assortment $\mathcal{A}=\mathcal{I}$.
A train dataset $\mathcal{D}_{train}$ and a test dataset $\mathcal{D}_{test}$, illustrated in Figure \ref{fig:synth_data}, are constituted by sampling baskets for the different following assortments:
\begin{itemize}
    \item Training set $\mathcal{D}_{train}$:
    \begin{itemize}
        \item Store 1: $\mathcal{A}_1$ = $\{1, 2, 4, 5, 7, 8\}= \mathcal{I} \backslash \{3, 6\}$ yields $\mathcal{D}_1$;
        \item Store 2: $\mathcal{A}_2$ = $\{1, 3, 5, 6, 7, 8\}= \mathcal{I} \backslash \{2, 4\}$ yields $\mathcal{D}_2$;
        \item Store 3: $\mathcal{A}_3$ = $\{2, 3, 4, 6, 7, 8\}= \mathcal{I} \backslash\{1, 5\}$ yields $\mathcal{D}_3$;
    \end{itemize}
    \item Test set $\mathcal{D}_{test}$:
    \begin{itemize}
        \item Store 4: $\mathcal{A}_4$ = $\{1, 2, 5, 6, 7, 8\}= \mathcal{I} \backslash \{3, 4\}$ yields $\mathcal{D}_4$;
        \item Full Assortment: $\mathcal{A}_{full} = \mathcal{I}$, i.e. all items  yields $\mathcal{D}_{full}$.
    \end{itemize}
\end{itemize}

\subsubsection{Protocol}

We compare several versions of our approach with other state of the art models: 
\begin{itemize}
    \item Shopper \cite{ruiz2020shopper}: A choice model that iteratively builds the most probable basket. It models the probability to add an item to an already partially built basket with price and interactions effects. The main difference lies in the expression of the basket probability that is computed by summing over all the possible ordered permutations of the basket.
    \item Basket-MVL \cite{basket2022caputo}: This choice model assumes the utility of a basket $\mathcal{B}$ to be the additive contributions of the marginal linear utilities of each item within the basket along with a symmetrical interaction term between those items: $U(\mathcal{B}) = \sum_{i \in \mathcal{B}} \beta \log p_i + \sum_{i, j \in \mathcal{B}} \boldsymbol{\gamma_i} \cdot \boldsymbol{\gamma_j}$. Finally, the probability for $\mathcal{B}$ to be purchased is computed with the softmax function over all possible baskets.
    \item Translation-Based Recommendation (TBR) \cite{translation}: It models a customer next choice as a translation from its previous choices with learned item-item interactions.
     \item $\text{Ours}$: Our model trained on $\mathcal{D}_{train}$ with negative samples drawn following Equation \ref{eq:NS}.
    \item $\text{Ours}_{\mathcal{I}}$: Our model trained on $\mathcal{D}_{train}$, however any item can be a negative sample as long as it is not in the basket. For this model, $\mathcal{NS} = \mathcal{I}\backslash \mathcal{B}_k$.
    \item $\text{Ours}_{full}$: Our model, trained on $\mathcal{D}_{full}$. This idealistic version of the model has been trained on the true underlying distribution. The two negative sampling strategies are the same for this model since $\mathcal{A} = \mathcal{I}$ in $\mathcal{D}_{full}$.
\end{itemize}

\begin{figure}
    \centering
    \includegraphics[width=1.0\linewidth]{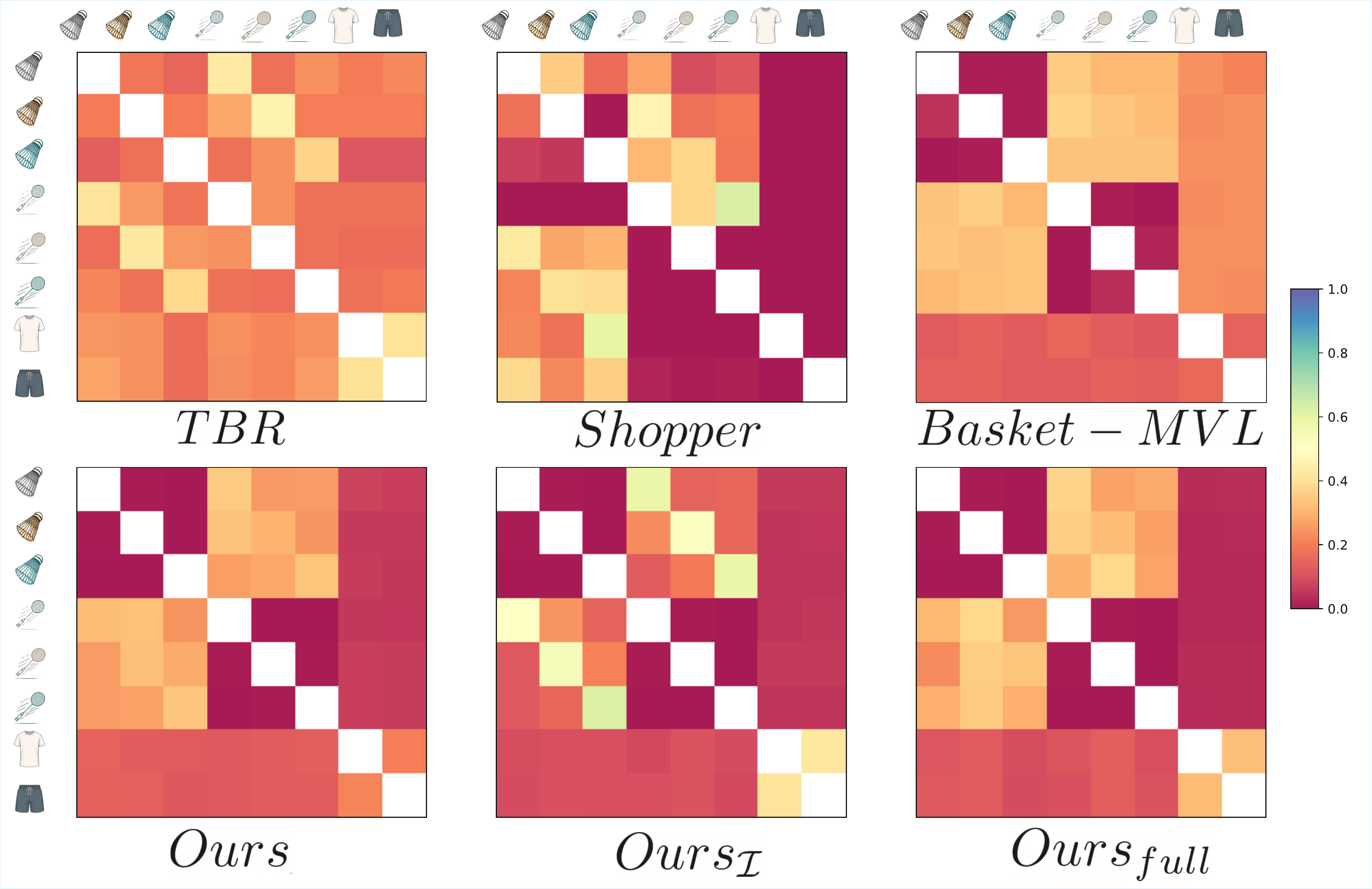}
    \caption{Conditional probabilities $\mathbb{P}(i|\mathcal{B}=\{j\})$ for $i \neq j$ as estimated by the different models on $\mathcal{D}_{full}$. These probabilities are to be compared with the left of Figure \ref{fig:interaction_matrix}.}
    \label{fig:afull_matrix}
\end{figure}

\begin{figure}
    \centering
    \includegraphics[width=1.0\linewidth]{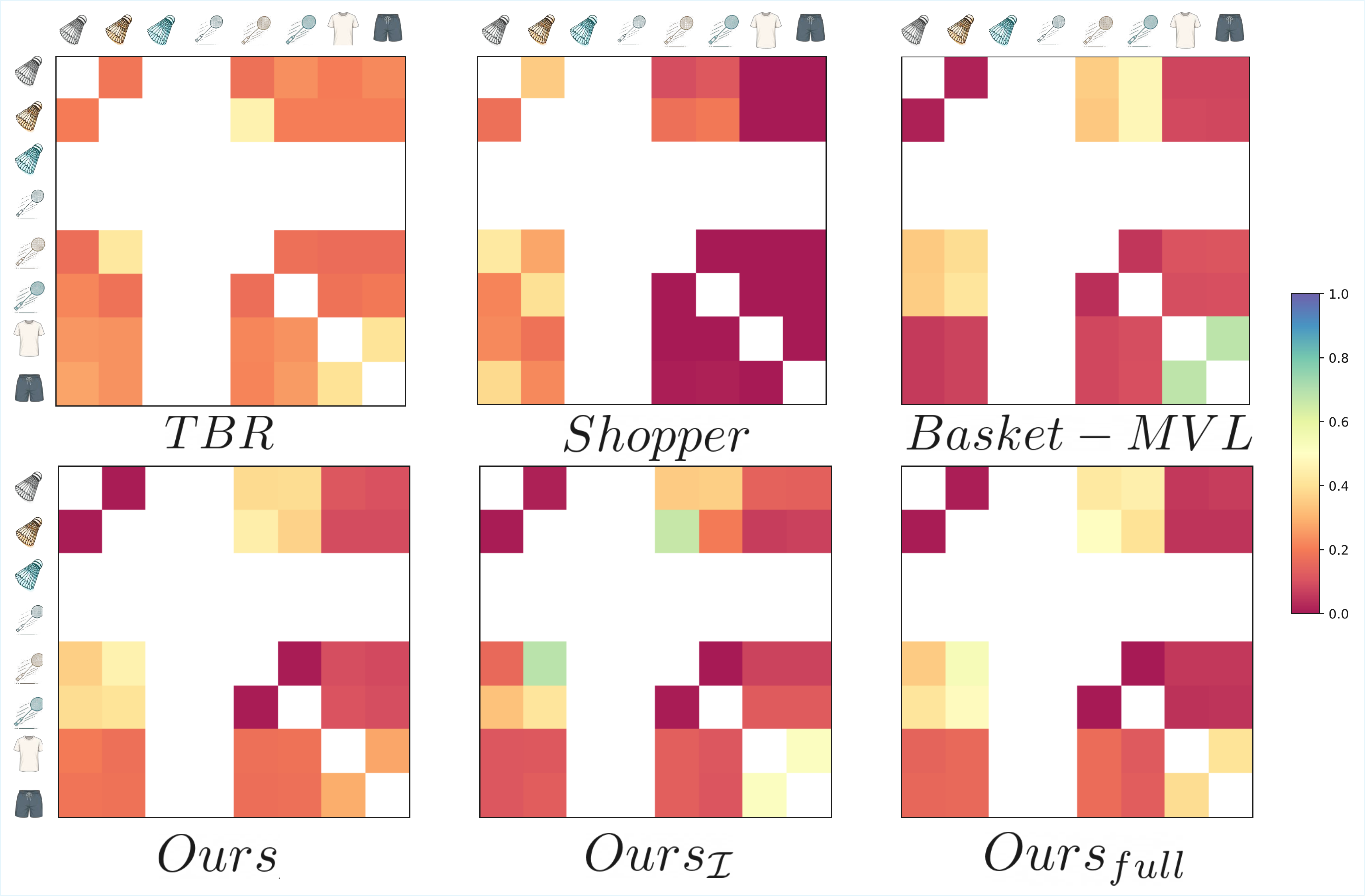}
    \caption{Conditional probabilities $\mathbb{P}(i|\mathcal{B}=\{j\})$ for $i \neq j$ as estimated by the different models on the test data $\mathcal{D}_{4}$. These probabilities are to be compared with the right of Figure \ref{fig:interaction_matrix}.}
    \label{fig:a4_matrix}
\end{figure}

Except for $\text{Ours}_{full}$, all models are trained with basket data $\mathcal{D}_{train}$ issued from the three assortment $\mathcal{A}_1, \mathcal{A}_2$ and $\mathcal{A}_3$, and evaluated on the test dataset $\mathcal{D}_{test}$. 
The differences between the models can be easily analyzed by comparing the co-purchase probabilities $\mathbb{P}(i \mid \mathcal{B}=\{j\})$ in Figure \ref{fig:afull_matrix} with the ground truth on $\mathcal{D}_{full}$ in Figure \ref{fig:interaction_matrix}. Similarly, Figure \ref{fig:a4_matrix} presents a new product combination with assortment $\mathcal{A}_4$, which is not part of the training set. In both cases, we observe that our model, $Ours_{\mathcal{A}}$ better captures the product interactions. We also evaluate the models with two metrics, the conditional Negative Log-Likelihood $NLL$ and the Mean Reciprocal Rank $MRR$ computed as:
\begin{equation}
    NLL(\mathcal{D}) = - \sum_{\mathcal{B} \in \mathcal{D}} \sum_{i \in \mathcal{B}} \log \mathbb{P}(i \mid \mathcal{B}\backslash i)
    \label{eq:test_nll}
\end{equation}
\begin{equation}
    MRR(\mathcal{D}) = \sum_{\mathcal{B} \in \mathcal{D}} \sum_{i \in \mathcal{B}}  \frac{1}{rank\left[V(i \mid \mathcal{B}\backslash i)\right]}
    \label{eq:test_mrr}
\end{equation}
with $rank[.]$ being the ranking function over the available options.

\begin{table}[t]
    \centering
    \caption{Negative Log-Likelihood $NLL$ (lower is better) and Mean Reciprocal Rank $MRR$ (higher is better) for the different models on the test set.}
    \setlength\tabcolsep{5.5pt}
    \begin{tabular}{lcccccc}
    \toprule
    Model & Shopper & Basket-MVL & TBR  & $\text{Ours}_{full}$ &  $\text{Ours}_{\mathcal{I}}$ & $\text{Ours}$ \\
    \midrule
     $NLL(\mathcal{D}_{full})$ & 1.75 & 1.69 & 1.82 & \textbf{1.25} & 1.43 & 1.26 \\
     $NLL(\mathcal{D}_{4})$ & 1.49 & 1.41 & 1.55 & 0.97& 1.07 & \textbf{0.95} \\
     $MRR(\mathcal{D}_{full})$& 0.32 & 0.41 & 0.28 & \textbf{0.42}& 0.39 & 0.41  \\
     $MRR(\mathcal{D}_{4})$& 0.45 & 0.60 & 0.45 & \textbf{0.62} & 0.59 & 0.61 \\
    \bottomrule
    \end{tabular}
    \label{tab:tab_1}
\end{table}

The metrics, reported in Table \ref{tab:tab_1}, show that taking well into account the assortment while drawing the negative samples is key to capture the proper patterns. Our proposition of negative sampling $\text{Ours}_{\mathcal{A}}$ better learns the true underlying distribution, and better generalizes on new assortment variations. A first explanation can be found using the conditional probabilities illustrated in Figures \ref{fig:afull_matrix}. Indeed, the difference is particularly visible for some pairs of complementary items, such as items $(1,6)$ and $(2,6)$ that have not always been sold at the same time. The models resulting from the naive negative sampling tend to underestimate the complementarity between such items (or equivalently overestimate cannibalization). 
Except few surprising outliers with high values, Shopper appears to worsen this effect, that we interpret as an impact of its encoding-decoding structure as well as the use of all sub-baskets during its training.
Taking into account the assortment is thus beneficial to capture well the different product interactions. While this notion might not be necessary for settings such as language \cite{mikolov2013distributed-rep-words} or online retail \cite{ruiz2020shopper}, it definitely should not be overlooked for brick-and-mortar retailers.

\subsection{Large scale experiments: Computation time and memory usage}
Computation time and memory usage are critical when dealing with basket choice models, particularly in real-world applications where models need to process large-scale sales data efficiently. While some open-source packages offer optimized processes for data storage and discrete choice models estimation \cite{choicelearn2024auriau,torchchoice2023du}, it is still a challenge when considering bundles of products. 

For this experiment, we compare the processing time for a single epoch of training with a dataset of one million purchases, with respectively $100$ and $1,000$ items considered. 
We compare our approach implemented in Python with TensorFlow \cite{abadi2016tensorflow} with the two other models. Shopper \cite{ruiz2020shopper}, has a C++ codebase publicly available. Additionally, we have re-implemented in Python a deterministic version of the model\footnote[3]{The code for the experiments can be found at \url{https://github.com/artefactory/alea-carta-est}.}, reproducing the training procedure for a non-Bayesian neural network. We also use our own Python implementation for the basket-MVL model, whose estimation uses a composite conditional likelihood function described in \cite{basket2022caputo}, and the Translation-Based Recommendation \cite{translation}. 
The experiment is conducted on a single AMD Ryzen PRO 5955WX CPU and 250GB of RAM.

\begin{table}[t]
    \centering
    \caption{Comparison of computation time and RAM usage for different models}\label{tab:time}
    \setlength\tabcolsep{5.5pt}
    \begin{tabular}{lcccc}
    \toprule
    & \multicolumn{2}{c}{100 Items} &  \multicolumn{2}{c}{1000 Items} \\
     & Process Time & Fits in RAM &  Process Time & Fits in RAM \\
    \midrule
     Shopper & 100s  & \checkmark & - & $\times$\\
     Python-Shopper & 110s & $\checkmark$ & 620s & $\checkmark$\\
     Basket-MVL & 3,000s & $\checkmark$  & - & $\times$ \\
     TBR & 550s & $\checkmark$  & 1020s & $\checkmark$ \\
     $\text{Ours}$ & 80 s & \checkmark & 360s & \checkmark\\
    \bottomrule
    \end{tabular}
\end{table}

The results are reported in Table \ref{tab:time}. First, we can observe that our approach as well as our Python implementation of Shopper scale way better in terms of dataset size, while it is competitive with the C++ implementation in terms of training speed. Our approach is also faster than the Shopper re-implementation, mainly because it does not need to consider permutations of the basket. 
Following these results, we discarded Shopper and Basket-MVL models for our large scale real-world dataset, with experiments described below.

\subsection{Experiments on real-world retail data}

We consider a dataset of purchases collected through hundreds of different supermarkets of the European retail company \textit{Fortenova Group} during four years, from 2017 to 2021 that we describe in Table \ref{tab:magnitude}. Restraining the dataset to food related products that have been sold at least 1,000 times, we count more than 6,000 different products purchased.
We consider a train/test temporal split to evaluate our model. The train data covers purchases from 2017 to 2020 and test data $\mathcal{D}_{2021}$ covers year 2021. In Table \ref{tab:fng_test_nll} we share some metrics results, stressing again the need for a negative sampling adapted to the assortment. The results were obtained with the same latent size $K=128$ for all embeddings and $|\mathcal{NS}| = 20$ negative samples. We share more details of the implementation in Appendix B.

\begin{table}[t]
    \centering
    \caption{Description of our retail dataset. Numbers are rounded for confidentiality reasons.}\label{tab:magnitude}
    \setlength\tabcolsep{5.5pt}
    \begin{tabular}{ccccc}
    \toprule
    \#Referenced Items &  \#Shops & \#Baskets & \#Assortments & Avg Basket Size  \\
    \midrule
     $6,000+$ & $600+$ & $100,000,000+$ & $10,000+$ & $12.4$ \\
    \bottomrule
    \end{tabular}
\end{table}

\begin{figure}[t]
\centering
\includegraphics[width=1.0\textwidth]{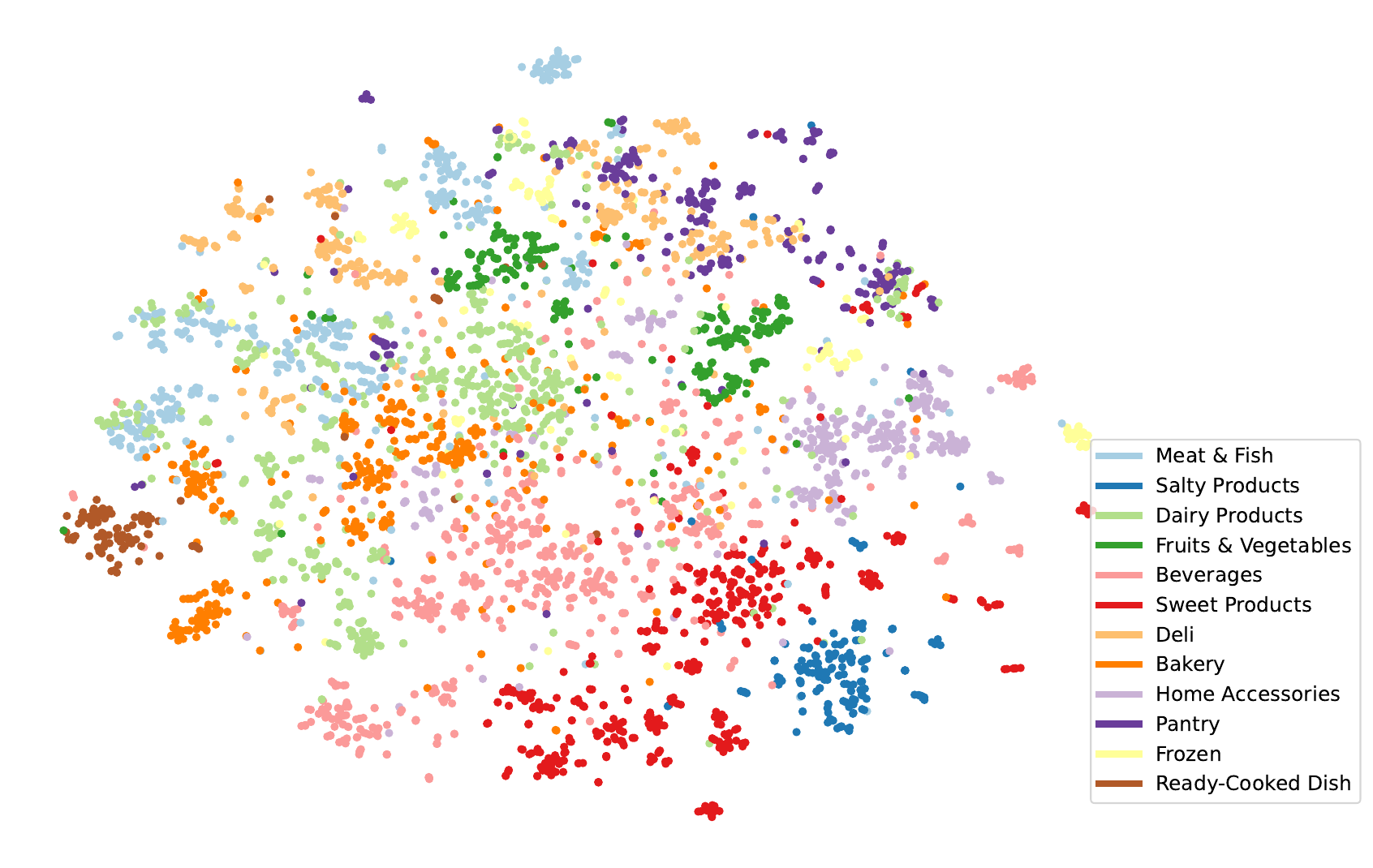}
\caption{T-SNE projection of the learned $\boldsymbol{\gamma}_i$ embeddings} \label{fig:tsne}
\end{figure}

In Figure \ref{fig:tsne}, we provide a T-SNE \cite{van2008visualizing} projection of a subset of the learned $\boldsymbol{\gamma}_i$ embeddings of our model, colored according to product $i$'s category. We observe that while the model is not provided with any type of information about the different products beside their price, it successfully creates clusters that correspond to products categories. It showcases that the model is able to learn a logical hierarchy of the products using the basket information. 

\begin{table}[t]
    \centering
    \caption{Conditional Negative Log-Likelihood (lower is better) and Mean Reciprocal Rank (higher is better) on the retail test set.}\label{tab:test_nll}
    \setlength\tabcolsep{5.5pt}
    \begin{tabular}{lcc}
    \toprule
    Model &  $\text{Ours}_{\mathcal{I}}$ & $\text{Ours}$  \\
    \midrule
     $NLL(\mathcal{D}_{2021})$ & 3.03 & \textbf{2.48} \\
     $MRR(\mathcal{D}_{2021})$ & 0.015 & \textbf{0.018 }\\
    \bottomrule
    \end{tabular}
    \label{tab:fng_test_nll}
\end{table}

\subsection{Application: Highlighting complementary products}

The main purpose of such models is to assist a category manager when planning the future assortment or promotion campaign for different stores. One basic experiment is to infer for each product a short list of the most complementary items. With such knowledge provided, the category manager can decide to make sure that complementary items are integrated within the same assortments. 
We propose to measure the following quantity $C(i,j) \in \mathbb{R}$ in order to rank the items $j$ according to their complementarity with item $i$:

\begin{equation}
    C(i, j) = \sum_{\mathcal{B} \in \mathcal{D}} \mathbb{P}(i\mid \mathcal{B} \cup \{j\}) + \mathbb{P}(j\mid \mathcal{B} \cup \{i\}) - \mathbb{P}(i\mid \mathcal{B} \backslash \{j\}) - \mathbb{P}(j\mid \mathcal{B} \backslash \{i\})
    \label{eq:complem}
\end{equation}
which is symmetrical w.r.t $i$ and $j$. The top $10$ products $j$ in terms of $C(i,j)$ values are shown in Table \ref{tab:compl2}, for $4$ examples of products $i$. In the case of tomato sauce and vanilla powder, the model outlines products that do appear highly complementary. We obtain respectively pasta, bread and pizza dough with tomato sauce and diverse ingredients to bake a cake with the vanilla powder, thus recovering meaningful ingredients together. The example of the soda bottles is also particularly interesting. Indeed, the model has understood packaging size interactions. The large soda bottle has other large bottles as complementary items, while the small soda bottles has smaller formats as top complementary products and also a single portion, to-go ice cream. It may correspond to two types of purchasing behaviors, one optimizing the price/quantity ratio with larger packages and the other one corresponding to instant consumption.

\begin{table}[t]
\caption{Examples of top-10 complementary products according to our model, ranked w.r.t $C(i,j)$ (Equation \ref{eq:complem}) for the product $i$ displayed on top. The products are referenced by their nature and brand names, replaced by a letter for anonymity. For each product displayed on top, the first-ranked complementary product is on the top left and the tenth-ranked complementary product is on the bottom right.}
\centering
\label{tab:compl2}
\setlength\tabcolsep{8.5pt}
\begin{tabular}{ll|ll}
\toprule
\multicolumn{2}{c}{Soda A \textbf{0.5L}}   & \multicolumn{2}{c}{Soda A \textbf{2L}}\\
\midrule
\small{Water A \textbf{0.5L}} & \small{Water B \textbf{1L}} &
\small{Soda B \textbf{2L}} & \small{Soda C \textbf{2L}} \\
\small{Soda B \textbf{0.5L}} & \small{Water C \textbf{0.5L}} & \small{Water A \textbf{1.5L}} & \small{Sweets A \textbf{500g}}  \\
\small{Soda C \textbf{0.5L}} & \small{Energy Drink \textbf{0.5L}} \hspace{-15pt} & \small{Soda D \textbf{1.5L}} & \small{Water B \textbf{1L}}\\
\small{Soda D \textbf{0.5L}} & \small{Strong Alcohol \textbf{0.1L}} \hspace{-10pt} & \small{Light Alcohol \textbf{0.5L}} \hspace{-20pt} & \small{Soda E \textbf{2L}}\\
\small{Water D \textbf{0.5L}} & \small{Ice Cream \textbf{0.16L}} & \small{Sweets B \textbf{100g}} & \small{Soda F \textbf{2L}}\\
\toprule
\multicolumn{2}{c}{Tomato sauce A}   & \multicolumn{2}{c}{Vanilla powder A}\\
\midrule
Pasta A & Pasta B &
Margarin A & Chocolate A \\
Bread A & Pizza Dough A& Chocolate B & Chocolate C\\
Pasta C & Other Sauce A & Baking Powder A  & Sugar A\\
Toasts A & Other Sauce B & Baking Powder B & Margarin B\\
 Bread B & Other Sauce C  & Whipped Cream A & Margarin C \hspace{-15pt}\\
\bottomrule
\end{tabular}
\end{table}

Following these promising results, we are currently conducting a survey with the company's category managers in order to confirm the relevancy of our top complementary items. They are asked to give feedback on different lists of items inferred by our model. Following these feedbacks, we plan to deploy the model as an API that would be called by several user interfaces used by category managers in their daily work. The model would regularly be trained on the sales data in order to keep up-to-date patterns.

\section{Conclusion}


In this work, we introduced a machine learning model for basket choice modeling that learns  embeddings of items in order to capture interactions between them. By leveraging negative sampling techniques, we addressed the computational challenges posed by large transaction datasets, ensuring efficient and scalable learning. Our experiments on synthetic data demonstrated the added value of our model and of its negative sampling method that better captures interactions between products in retail. The large-scale evaluation on more than one million transactions from a leading European retailer validated the practical applicability of our approach, with notably very promising top complementary items. 
Besides, our approach shows competitive predictive performances and computing scalability, in particular when compared to the models from the literature.

The results confirm the potential of our model to enhance retail decision-making, particularly in helping category managers with assortment optimization and promotion campaign planning. Beyond the direct applications by enriching user interfaces in retail, our model can be extended to other domains involving large-scale consumer choice modeling. Future work will explore the integration of additional features, such as spatial proximity between products in the store shelves and personalized customer profiles. The former will enable the identification of product interactions with respect to their distance in store, while the latter will reveal how these interactions vary across different customer segments.

\bibliographystyle{plain}

\newpage

\appendix

\input{appendix_a}
\input{appendix_b}

\end{document}

%% file: appendix_a.tex
\section{Synthetic Data Generation}
\label{app:sampling}
We describe the iterative algorithm used for the experiments in Section \ref{sec:synthetic} to obtain synthetic data. From an assortment $\mathcal{A}$, we sample the purchased baskets as follows:

\textbf{1. Selection of the first item}
We consider that the following sets can be chosen with the same probability:
\begin{itemize}
    \item $\{1, 2, 3\} \cap \mathcal{A}$ the ensemble of available shuttlecocks
    \item $\{4, 5, 6\} \cap \mathcal{A}$ the ensemble of available rackets
    \item $\{7\} \cap \mathcal{A}$ the ensemble of available shorts
    \item $\{8\} \cap \mathcal{A}$ the ensemble of available t-shirts
\end{itemize}
One item is then chosen among the selected set, all items also having the same probability.

\textbf{2. Basket Completion}
Once the first item $i$ is chosen, one item by remaining nest can be added to the basket.
Probabilities $\mathbb{P}(i | j \in \mathcal{B})$  equal to $0.7$ for nest with complementary items and $0.15$ for neutral items. Once a nest is selected, one items among the nest is finally selected, all having the same probability.

\textbf{3. Adding some random noise}
With a probability $P = 0.15 $ one of the items of the assortment, not already in the basket, may also be added.

\begin{figure}[h]
\centerline{\includegraphics[width=1.0\textwidth, keepaspectratio]{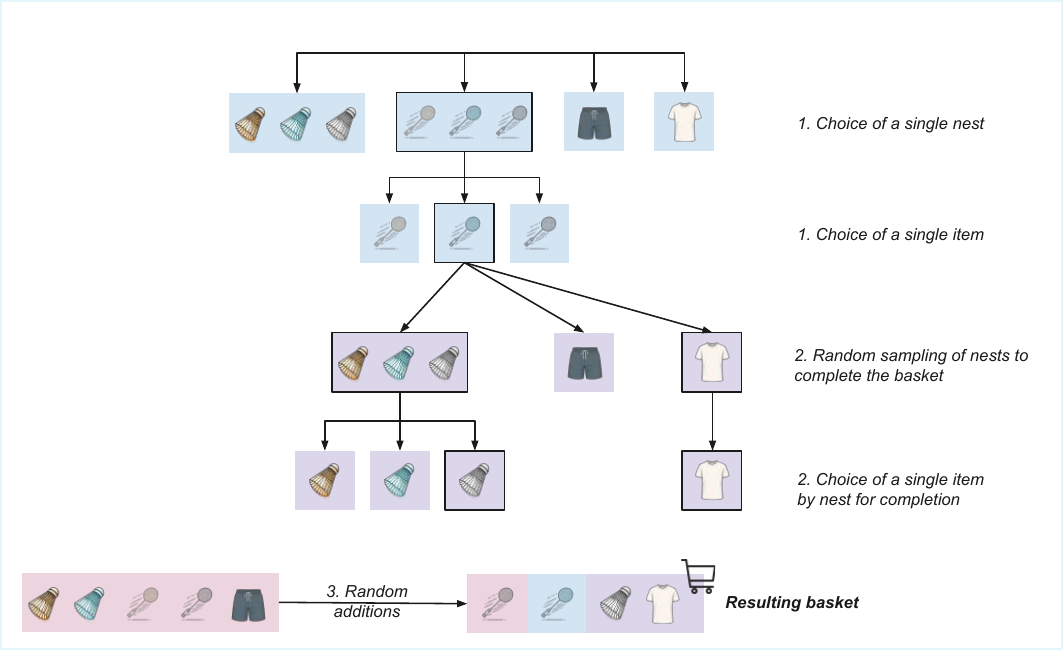}}
\caption{Successive steps for synthetic data sampling from an assortment. In this example, the racket nest is selected during the first step, with the green racket finally being selected. During the step 2, the grey shuttlecock and the t-shirt are added. Finally a second racket is added to the basket during step 3.}\label{fig:synth_sampling}
\end{figure}

%% file: appendix_b.tex
\section{Experiments on real-world retail data}

\label{app:paramaters}
For the experiments on real-world data, we used the following hyperparameters:

\begin{itemize}
    \item $\alpha$, $\beta$ latent size: 12
    \item $\gamma$ latent size: 128
    \item Number of negative samples: $|\mathcal{NS}| = 24$
    \item Optimizer: Adamax with a learning rate of 0.001
    \item Batch size: 256
    \item Number of epochs: 20
\end{itemize}